\documentclass[runningheads,a4paper]{llncs}

\usepackage{amssymb}
\usepackage{graphicx}
\usepackage{mathrsfs}
\usepackage{amsmath}
\usepackage[all]{xy}
\usepackage{psfrag}

\usepackage{url}
\urldef{\mailsa}\path|{dab54,pao32}@cam.ac.uk|

\newcommand{\field}[1]{\mathbb{#1}}

\newcommand{\fs}[1]{\mathcal{#1}}

\newcommand{\reals}{\field{R}}
\newcommand{\define}{:=}
\newcommand{\comment}[1]{}

\newcommand{\defterm}[1]{\textbf{#1}}

\newcommand{\prob}{\mathbf{P\!r}}   
\newcommand{\agent}{\mathbf{P}}     
\newcommand{\utility}{\mathbf{U}}   
\newcommand{\ug}{\mathbf{u}}   

\newcommand{\energy}{\mathbf{e}}    
\newcommand{\fu}{\mathbf{J}}        

\newcommand{\expect}{\mathbf{E}}    

\begin{document}

\mainmatter  

\title{Information, Utility \& Bounded Rationality\thanks{A shortened version of this paper has been published in Lecture Notes on Artificial Intelligence 6830, pp. 269--274.}}

\titlerunning{Bounded Rationality}

%
%
\author{Pedro A. Ortega \and Daniel A. Braun}
\authorrunning{Bounded Rationality}

\institute{Department of Engineering, University of Cambridge\\
Trumpington Street, Cambridge, CB2 1PZ, UK\\
\mailsa\\
}

%
%

\toctitle{Lecture Notes in Artificial Intelligence}
\tocauthor{Bounded Rationality}
\maketitle

\begin{abstract}
Perfectly rational decision-makers maximize expected utility, but crucially ignore the resource costs incurred when determining optimal actions. Here we propose an axiomatic framework for bounded rational decision-making based on a thermodynamic interpretation of resource costs as information costs. We show that
this axiomatic framework enforces a unique conversion law between utility and information, which can be characterized by a variational ``free utility'' principle akin to thermodynamical free energy. This variational principle constitutes a normative criterion that trades off utility and information costs, the latter measured by the Kullback-Leibler deviation between a distribution representing a desired policy and a reference distribution representing an initial default policy. We show that bounded optimal control solutions can be derived from this variational principle, which leads in general to stochastic policies.
Furthermore, we show that risk-sensitive and robust (minimax) control schemes fall out naturally from this framework if the environment is considered as an adversarial opponent. When resource costs are ignored, the maximum expected utility principle is recovered.

\end{abstract}

\section{Introduction}

Rational decision-making is usually based on the principle of
maximum expected utility (MEU) \cite{Neumann1944}.
According to MEU, a rational agent chooses its action $a$ so as
to maximize its expected utility
$\expect[\utility|a] = \sum_s \agent(s|a) \utility(s)$
given the probability $\agent(s|a)$ that action $a \in \fs{A}$ will lead to
outcome $s \in \fs{S}$ and given that the desirability of the outcome $s$ is
measured by the utility $\utility(s) \in \reals$. Thus, expected utilities
express betting preferences over lotteries with uncertain outcomes. The optimal
action $a^\ast \in \fs{A}$ is defined as the one that maximizes the expected
utility, that is $a^\ast \define \arg \max_a \expect[\utility|a]$.
However, computing such
optimal actions is often very difficult in practice due to prohibitive resource costs
that are associated with the process of finding the optimal action. Such resource costs
are ignored by MEU.

In contrast, a bounded rational decision-maker has only limited resources
and cannot afford an
unlimited search for the optimal action \cite{Simon1982}. Therefore, such decision-makers
have to trade off the utility that an action achieves against the resource cost of finding the action. Imagine, for example, you want to invest some of your savings and you start reading up on several options, asking your local bank, etc. However, as a bounded agent
in the real world you cannot extend this search forever, as you will loose out in the
meanwhile. Therefore, you have to trade off somehow the time invested in this search and a satisfactory return from some investment option.

In this paper we propose an axiomatic formalization of bounded rationality that leads
to such a trade-off based on a thermodynamic interpretation of resource costs \cite{Feynman1996}. The intuition behind this interpretation is that ultimately any real decision-maker has to be incarnated in a physical system, since any process of information processing must always be accompanied by a pertinent physical process \cite{Tribus1971}.
Thermodynamics provides the tools to study these general physical systems. In Section~2
we discuss the thermodynamical notion of resource costs in information processing systems. In Section~3 we show how a set of simple choice axioms leads to a variational principle
that allows computing bounded optimal policies in systems with resource costs. In Section~4 we apply this framework and show how to derive bounded optimal solutions for decision-making under resource costs in different environments. We also show how to obtain classic maximum expected utility solutions in the limit of negligible resource costs.

\section{Resource Costs}

In the following we conceive of information processing as changes in information states, i.e. ultimately changes of probability distributions that are represented in physical systems.
Changing an information state therefore implies changes in physical states, such as flipping gates in a transistor, changing voltage on a microchip, or even changing location of a gas particle. Changing such states is costly and requires thermodynamical work \cite{Feynman1996}.
Imagine, for example, that we use an ideal gas particle in a box with volume $V_i$ as an
information processing system to
represent a uniform probability density over a random variable with $p_i = \frac{1}{V_i}$. If we now want to update this probability to $p_f$, because we gained information $- \log p= -\log \frac{p_i}{p_f} > 0$, we have to
reduce the original volume to $V_f = p V_i$. However,
this decrease in volume requires the work $W = - \int_{V_i}^{V_f} \frac{N k T}{V} dV = N k T \ln \frac{V_i}{V_f}$, where $N$ is the number of gas molecules, $k$ is the Boltzmann constant, and $T$ is temperature. Thus, in this simple example we can compute the relation between the change in information state
and the required work, that is $W=-\alpha \log p$, with $\alpha = \frac{k T}{\log e} > 0$ being the conversion factor between information and energy. The conversion factor $\alpha$ depends on the underlying
  properties of the physical system and determines how
  expensive it is to process information. In the next two sections, we derive
a general expression of information costs for physical systems that represent
bounded rational decision-makers. Since such decision-makers need to trade off utility and information costs, we will first investigate the relation between information and utility \cite{OrtegaBraun2010a} and then show how information costs appear as an additional term in the utility in physically implemented decision-makers.

\section{Conversion between utility and information}

\subsection{Choice axioms}\label{sec:utility}

Consider a decision-maker whose behavior is represented by a probability space $(\Omega, \fs{F}, \agent)$ with sample set $\Omega$ and $\sigma$-algebra $\fs{F}$ of measurable events
between which the decision-maker can choose. We assume that the decision-maker can \emph{choose freely} any probability measure $\agent$ representing his choice behavior. Thus, if $\agent(A) > \agent(B)$, then the propensity of choosing $A$ is higher than that of choosing $B$. This difference in probability can be given a \emph{utilitarian} interpretation: \emph{$A$ is chosen with higher probability than $B$ because $A$ is more desirable than $B$.}
The measure that quantifies such differences in desirability is commonly called a \defterm{utility} function. If there is such a measure, then it is reasonable to demand the following properties:
\begin{itemize}
    \item[i.] Utilities should be mappings from events into real numbers.
    \item[ii.] Absolute values of utility are irrelevant, only relative differences in utility should matter (``utility gains'').
    \item[iii.] Utility gains should be additive.
    \item[iv.] A decision-maker should assign more probability mass to events with high utility and less probability mass to events with low utility.
    \item[v.] An adversarial agent should make the reverse assignment of probability mass.
\end{itemize}
These postulates are summarized in the following definition.

\begin{definition}[Axioms of Choice]\label{def:utility}
Let $(\Omega, \fs{F}, \agent)$ be a probability space. A set function
$\utility: \fs{F} \rightarrow \reals$ is a \defterm{utility function} for
a decision-maker with probability measure
$\agent$ iff its \defterm{utility gain function} $\ug(A|B) \define
\utility(A \cap B) - \utility(B)$ has the following three properties for all
events $A, B, C, D \in \fs{F}$:
\begin{align*}
    \text{A1.}\quad
        & \exists f, \ug(A|B) = f\bigr( \agent(A|B) \bigr)\in \reals,
        && \text{(real-valued)}\\
    \text{A2.}\quad
        & \ug(A \cap B|C) = \ug(A|C) + \ug(B|A \cap C),
        && \text{(additive)}\\
    \text{A3.}\quad
        & \agent(A|B) > \agent(C|D)
            \quad \Leftrightarrow \quad
            \ug(A|B) > \ug(C|D).
        && \text{(monotonic increasing)}
\end{align*}
If the decision-maker is an adversarial opponent, the inequality of A3 is reversed
\begin{align*}
    \text{A4.}\quad
        & \agent(A|B) > \agent(C|D)
            \quad \Leftrightarrow \quad
            \ug(A|B) < \ug(C|D).
        && \text{(monotonic decreasing)}
\end{align*}
Furthermore, we use the abbreviation $\ug(A) \define \ug(A|\Omega)$.
\end{definition}

The following theorem shows that these three properties enforce a strict
mapping between probabilities and utility gains.

\begin{theorem}[Utility Gain $\leftrightarrow$ Probability]\label{theo:utility}
If $f$ is such that $\ug(A|B) = f(\agent(A|B))$ for any probability space
$(\Omega, \fs{F}, \agent)$, then $f$ is of the form
\[
    f(\cdot) = \alpha \log(\cdot),
\]
where $\alpha$ is an arbitrary strictly positive constant in case of A3 or
an arbitrary strictly negative constant in case of A4.
\end{theorem}

The proof is provided elsewhere \cite{OrtegaBraun2010a,Ortega2011}.
If one is willing to accept Definition~\ref{def:utility}, then one obtains the relations
\begin{equation}\label{eq:cross-connections}
    \utility(A \cap B) - \utility(B)
    = \alpha \log \agent(A|B).
\end{equation}
In this relation, $\alpha$ plays the role of a conversion factor between
utilities and information. A bounded rational decision-maker is characterized by $\alpha>0$,
whereas an adversarial opponent can be described by $\alpha<0$. Unless otherwise stated,
we will assume $\alpha>0$ in the following. If a probability measure $\agent$ and a utility
function $\utility$ satisfy the relation \eqref{eq:cross-connections}, then we
say that they are \defterm{conjugate}. Given that this transformation between utility gains and
probabilities is a bijection, one can rewrite any probability $\agent(A|B)$ as a Gibbs
measure:
\begin{equation}
\label{eq:gibbs}
    \agent(A|B)
    = \frac{ \sum_{\omega \in A \cap B} \exp^{\frac{1}{\alpha} \utility(\omega)} }
           { \sum_{\omega \in B} \exp^{\frac{1}{\alpha} \utility(\omega)} }.
\end{equation}
where we have used the abbreviation $\utility(\omega) \define
\utility(\{\omega\})$. This transformation implies that the probability measure $\agent$ is
the Gibbs measure with temperature $\alpha$ and energy levels $\energy(\omega)
\define -\utility(\{\omega\})$. As the conversion factor $\alpha$ approaches zero, the
probability measure $\agent(\omega)$ approaches a delta function $\delta_{\omega^*}(\omega)$ with $\omega^* = \arg \max_{\omega} \utility(\omega)$, or in case of several maxima the uniform distribution over the
maximal set $\Omega_{\max} \define \{ \omega^\ast \in \Omega | \omega^\ast =
\arg \max_\omega \utility(\omega)\}$ .
Similarly, as $\alpha \rightarrow
\infty$, $\agent(\omega) \rightarrow \frac{1}{|\Omega|}$, i.e.\ the uniform
distribution over the whole outcome set $\Omega$.

\subsection{Variational principle}\label{sec:variational-principle}

It is well known in statistical physics that the Gibbs measure
satisfies a variational problem in the free energy \cite{Callen1985}.
Since utilities correspond to negative energies, we can formulate a
\emph{free utility} principle that is maximized by a decision-maker that acts
according to (\ref{eq:gibbs}).

\begin{theorem}
Let $X$ be a random variable with values in $\fs{X}$. Let $\agent$ and
$\utility$ be a conjugate pair of probability measure and utility function over
$X$. Define the \defterm{free utility} functional as
\[
    \fu(\prob; \utility) \define \sum_{x \in \fs{X}} \prob(x) \utility(x)
        - \alpha \sum_{x \in \fs{X}} \prob(x) \log \prob(x),
\]
where $\prob$ is an arbitrary probability measure over $X$. Then,
\[
    \fu(\prob; \utility) \leq \fu(\agent; \utility) = \utility(\Omega).
\]
\end{theorem}
A proof can be found in \cite{Keller1998}.
The free utility is a combined measure of a system's
expected utility and its uncertainty. The
variational principle implies that the Gibbs measure $\agent$ maximizes the free utility for a given utility function
$\utility$, as $\agent = \arg \max_\prob \fu(\prob; \utility)$.


The variational principle of the free utility also
allows measuring the cost of transforming the state of a stochastic
system required for information processing. 
Consider an initial system having
probability measure $\agent_i$ and utility function $\utility_i$. This system
satisfies the equation
\[
    \fu_i \define
    \sum_{x \in \fs{X}} \agent_i(x) \utility_i(x)
    - \alpha \sum_{x \in \fs{X}} \agent_i(x) \log \agent_i(x)
    = \utility_i(\Omega).
\]
If we add new constraints represented by the utility function $\utility_*$ then
the resulting utility function $\utility_f$ is given by the sum
\[
    \utility_f = \utility_i + \utility_*,
\]
and the resulting probability measure $\agent_f$ maximizes
\begin{align*}
    \fu(\prob, \utility_f)
    &= \sum_{x \in \fs{X}} \prob(x) \utility_f(x)
        - \alpha \sum_{x \in \fs{X}} \prob(x) \log \prob(x) \\
    &= \sum_{x \in \fs{X}} \prob(x) (\utility_i(x) + \utility_*(x))
        - \alpha \sum_{x \in \fs{X}} \prob(x) \log \prob(x) \\
    &= \sum_{x \in \fs{X}} \prob(x) \utility_*(x)
        - \alpha \sum_{x \in \fs{X}} \prob(x) \log \frac{ \prob(x) }{ \agent_i(x) }
        + \utility_i(\Omega).
\end{align*}
Let $\fu_f \define \fu(\agent_f, \utility_f)$. The difference in free utility
is
\begin{equation}\label{eq:free-utility-change}
    \fu_f - \fu_i
    = \sum_{x \in \fs{X}} \agent_f(x) \utility_*(x)
        - \alpha \sum_{x \in \fs{X}} \agent_f(x) \log \frac{ \agent_f(x) }{ \agent_i(x) }.
\end{equation}

In physical systems with constant $\alpha$, this difference measures the amount of work
necessary to change the state of the system from state $i$ to state $f$. The first term
of the equation measures the expected utility difference $\mathbf{U}_*(x)$, while the second term measures the information cost of transforming the probability distribution from state $i$ to state $f$. These two terms can be interpreted as determinants of
bounded rational decision-making in that they formalize a trade-off between an
expected utility $\mathbf{U}_*$ (first term) and the information cost of transforming $\mathbf{P}_i$ into $\mathbf{P}_f$ (second term). In this interpretation $\mathbf{P}_i$
represents an initial probability or policy, which includes the special case of the
uniform distribution where the decision-maker has initially no preferences. Deviations from
this initial probability incur an information cost measured by the KL divergence. If this
deviation is bounded by a non-zero value, we have a bounded rational agent.

In thermodynamics there are two dominant formulations of the second law that allow determining the equilibrium distribution:
the first and maybe more familiar formulation is the principle of maximum entropy, and the second principle is the principle of minimum energy \cite{Callen1985}. The corresponding variational
problems are typically formulated such that in the case of maximum entropy we hold the
mean energy fixed (i.e. in our case the expected utility), and in the case of minimum energy
(i.e. in our case maximum utility) we hold the entropy fixed. Mathematically, the constraints of fixed entropy and fixed utility are added by Lagrange multipliers. In our context
with respect to equation~\ref{eq:free-utility-change} this leads to two different variational principles:

\begin{enumerate}
    \item \textbf{Control.} The minimum energy principle translates into a bounded maximum utility principle. Given an initial policy represented by the probability
        measure $\agent_i$ and the constraint utilities $\utility_*$, we are looking for the final system $\agent_f$ that optimizes
        the trade-off between utility and resource costs. That is,
    \begin{equation}\label{eq:fu-endogenous}
    \agent_f = \arg \max_\prob
    \sum_{x \in \fs{X}} \prob(x) \utility_*(x)
        - \alpha \sum_{x \in \fs{X}} \prob(x) \log \frac{ \prob(x) }{ \agent_i(x) }.
    \end{equation}
    The solution is given by
    \[
        \agent_f(x) \propto \agent_i(x)
            \exp\biggl( \frac{1}{\alpha} \utility_*(x) \biggr).
    \]
    In particular, at very low temperature $\alpha \approx 0$,
    (\ref{eq:free-utility-change}) becomes
    \[
        \fu_f - \fu_i
        \approx \sum_{x \in \fs{X}} \agent_f(x) \utility_*(x),
    \]
    and hence resource costs are ignored in the choice of $\agent_f$, leading to
    $\agent_f \approx \delta_{x^\ast}(x)$, where $x^\ast = \max_x \utility_*(x)$. Similarly, at
    a high temperature, the difference is
    \[
        \fu_f - \fu_i
        \approx - \alpha \sum_{x \in \fs{X}} \agent_f(x)
            \log \frac{ \agent_f(x) }{ \agent_i(x) },
    \]
    and hence only resource costs matter, leading to $\agent_f \approx \agent_i$.
    \item \textbf{Estimation.} The maximum entropy principle translates into a minimum
    relative entropy principle for estimation. Given a final probability measure $\agent_f$
    that represents the environment and the constraint utilities $\utility_*$, we
    are looking for the initial system $\agent_i$ that satisfies
    \begin{align}
        \agent_i &= \arg \max_\prob
        \sum_{x \in \fs{X}} \agent_f(x) \utility_*(x)
            - \alpha \sum_{x \in \fs{X}} \agent_f(x) \log \frac{ \agent_f(x) }{ \prob(x) }
            \label{eq:fu-exogenous} \\
        &= \arg \min_\prob
        \sum_{x \in \fs{X}} \agent_f(x) \log \frac{ \agent_f(x) }{ \prob(x) },
            \nonumber
    \end{align}
    and thus we have recovered the minimum relative entropy principle for
    estimation, having the solution
    \[
        \agent_i = \agent_f.
    \]
\end{enumerate}
The minimum relative entropy principle for estimation
is well-known in the literature as it underlies Bayesian inference \cite{Opper1997}, but
the same principle can also be applied to problems of adaptive control \cite{OrtegaBraun2010b}.
In the following we focus on applications of the first principle on bounded optimal control.

\section{Applications}

Consider a system that first emits an action symbol $x_1$ with probability $P_0(x_1)$
and then expects a subsequent input signal $x_2$ with probability $P_0(x_2|x_1)$. Now we impose
a utility on this decision-maker that is given by $U(x_1)$ for the first symbol and
$U(x_2|x_1)$ for the second symbol. How should this system adjust its action probability $P(x_1)$ and expectation $P(x_2|x_1)$? Given the boundedness constraints $c_1$ and $c_2$ on the relative entropies,
the variational problem is given by
\begin{eqnarray}
\max_{p(x_1)p(x_2|x_1)} & & \sum_{x_1} p(x_1) U(x_1)
- \alpha \left( \sum_{x_1} p(x_1) \log \frac{p(x_1)}{p_0(x_1)} - c_1  \right)
+ \sum_{x_1,x_2} p(x_1) p(x_2|x_1) U(x_2|x_1) \nonumber \\
&-& \beta \left( \sum_{x_1,x_2} p(x_1) p(x_2|x_1) \log \frac{p(x_2|x_1)}{p_0(x_2|x_1)} - c_2 \right) , \nonumber
\end{eqnarray}
with $\alpha$ and $\beta$ as Lagrange multipliers.
We can rewrite this sum as a nested expression and drop all constants
\begin{eqnarray}
\max_{p(x_1)p(x_2|x_1)} \sum_{x_1} p(x_1) \Bigg[ U(x_1)
- \alpha \log \frac{p(x_1)}{p_0(x_1)}
+ \sum_{x_2} p(x_2|x_1) \bigg[U(x_2|x_1)
- \beta  \log \frac{p(x_2|x_1)}{p_0(x_2|x_1)} \bigg] \Bigg] . \nonumber
\end{eqnarray}
We have then an inner variational problem:
\begin{eqnarray}
\max_{p(x_2|x_1)} \sum_{x_2} p(x_2|x_1) \left[ -\beta \log \frac{p(x_2|x_1)}{p_0(x_2|x_1)} + U(x_2|x_1) \right]
\end{eqnarray}
with the solution
\begin{eqnarray}
p(x_2|x_1) = \frac{1}{Z_2} p_0(x_2|x_1) \exp\left( \frac{1}{\beta} U(x_2|x_1) \right)
\end{eqnarray}
and the $x_1$-dependent normalization constant
\[
Z_2 = \sum_{x_2} p_0(x_2|x_1) \exp\left( \frac{1}{\beta} U(x_2|x_1) \right)
\]
and an outer variational problem
\begin{eqnarray}
\max_{p(x_1)} \sum_{x_1} p(x_1) \left[ -\alpha \log \frac{p(x_1)}{p_0(x_1)} + U(x_1) + \beta \log Z_2 \right]
\end{eqnarray}
with the solution
\begin{eqnarray}
\label{eq:recursion}
p(x_1) &=& \frac{1}{Z_1} p_0(x_1) \exp\left( \frac{1}{\alpha} \left( U(x_1) + \beta \log Z_2 \right) \right)  \\
&=& \frac{1}{Z_1} p_0(x_1) \exp\left( \frac{1}{\alpha} \left( U(x_1) + \beta \log \sum_{x_2} p_0(x_2|x_1) \exp\left( \frac{1}{\beta} U(x_2|x_1) \right) \right) \right)\nonumber
\end{eqnarray}
and the normalization constant
\begin{eqnarray}
Z_1 &=& \sum_{x_1} p_0(x_1) \exp\left( \frac{1}{\alpha} \left( U(x_1) + \beta \log Z_2 \right) \right) \nonumber \\
&=& \sum_{x_1} p_0(x_1) \exp\left( \frac{1}{\alpha} \left( U(x_1) + \beta \log \sum_{x_2} p_0(x_2|x_1) \exp\left( \frac{1}{\beta} U(x_2|x_1) \right) \right) \right) \nonumber.
\end{eqnarray}
For notational convenience we introduce $\lambda=\frac{1}{\alpha}$ and $\mu=\frac{1}{\beta}$. Depending on the values of $\lambda$ and $\mu$ we can discern the following cases:

\begin{enumerate}

\item{\textbf{Risk-seeking bounded rational agent: }$\lambda>0$ and $\mu>0$}\\
When $\lambda>0$ the agent is bounded and acts in general stochastically.
When $\mu>0$ the agent considers the move of the environment as if it was his own
move (hence ``risk-seeking'' due to the overtly optimistic view). This follows
immediately from the choice axioms presented in section~\ref{sec:utility}.
We can also see this from the relationship between $Z_1$ and $Z_2$ in (\ref{eq:recursion}),
if we assume $\mu=\lambda$ and introduce the value function $V_t=\frac{1}{\lambda} \log Z_t$, which
results in the recursion
\[
V_{t-1} = \frac{1}{\lambda} \log \sum_{x_{t-1}} P_0(x_{t-1}|\cdot) \exp\left(\lambda \left(U(x_{t-1}|\cdot) + V_t \right)\right).
\]
Similar recursions based on the log-transform have been previously exploited for efficient approximations of optimal control solutions both in the discrete and the continuous domain \cite{Braun2011,Kappen2005,Todorov2009}.
In the perfectly rational limit $\lambda \rightarrow +\infty$, this recursion becomes the well-known Bellman recursion
\[
V^*_{t-1} = \max_{x_{t-1}} \left( U(x_{t-1}|\cdot) + V^*_t \right)
\]
with $V^*_t = \lim_{\lambda \rightarrow +\infty} V_t$.

\item{\textbf{Risk-neutral perfectly rational agent:} $\lambda \rightarrow +\infty$ and $\mu \rightarrow 0$ }\\
This is the limit for the standard optimal controller. We can see this from (\ref{eq:recursion}) by noting that
\[
\lim_{\mu \rightarrow 0} \frac{1}{\mu} \log \sum_{x_2} p_0(x_2|x_1) \exp\left( \mu U(x_2|x_1) \right) = \sum_{x_2} p_0(x_2|x_1) U(x_2|x_1),
\]
which is simply the expected utility. By setting $U(x_1) \equiv 0$, and taking the
limit $\lambda \rightarrow +\infty$ in (\ref{eq:recursion}), we therefore obtain
an expected utility maximizer
\[
p(x_1)=\delta(x_1-x_1^*)
\]
with
\[
x_1^* = \arg \max_{x_1} \sum_{x_2} p_0(x_2|x_1) U(x_2|x_1).
\]
As discussed previously, action selection becomes deterministic in the perfectly rational limit.

\item{\textbf{Risk-averse perfectly rational agent:} $\lambda \rightarrow +\infty$ and $\mu < 0$}\\
When $\mu<0$ the decision-maker assumes a pessimistic view with respect to the environment, as if the environment was an adversarial or malevolent agent. This attitude is sometimes called risk-aversion, because such agents act
particularly cautiously to avoid high uncertainty. We can see this from (\ref{eq:recursion}) by writing a Taylor series expansion for small $\mu$
\[
\frac{1}{\mu} \log \sum_{x_2} p_0(x_2|x_1) \exp\left( \mu U(x_2|x_1) \right) \approx \mathbb{E}[U] - \frac{\mu}{2} \mathbb{VAR}[U],
\]
where higher than second order cumulants have been neglected. The name risk-sensitivity then stems from the fact that variability or uncertainty in the utility of the Taylor series is subtracted from the expected utility.
This utility function is typically \emph{assumed} in risk-sensitive control
schemes in the literature \cite{Whittle1990}, whereas here it falls out naturally.  The perfectly rational actor with risk-sensitivity $\mu$ picks the action
\[
p(x_1)=\delta(x_1-x_1^*)
\]
with
\[
x_1^* = \arg \max_{x_1} \frac{1}{\mu} \log \sum_{x_2} p_0(x_2|x_1) \exp\left( \mu U(x_2|x_1) \right),
\]
which can be derived from (\ref{eq:recursion}) by setting $U(x_1) \equiv 0$ and by taking the limit $\lambda \rightarrow +\infty$. Within the framework proposed in this paper
we might also interpret the equations such that the decision-maker considers
the environment as an adversarial opponent with bounded rationality $\mu$.

\item{\textbf{Robust perfectly rational  agent:} $\lambda \rightarrow +\infty$ and $\mu \rightarrow -\infty$}\\
    When $\mu \rightarrow -\infty$ the decision-maker makes a worst case assumption about
    the adversarial environment, namely that it is also perfectly rational. This leads to the well-known game-theoretic minimax problem with the solution
\[
x_1^* = \arg \max_{x_1} \arg \min_{x_2} U(x_2|x_1),
\]
which can be derived from (\ref{eq:recursion}) by setting $U(x_1) \equiv 0$, taking the limits $\lambda \rightarrow +\infty$ and $\mu \rightarrow -\infty$ and by noting
that $p(x_1)=\delta(x_1-x_1^*)$. Minimax problems have been used to reformulate robust control problems that allow controllers to cope with model uncertainties \cite{Basar1995}. Robust control problems are also known to be related
to risk-sensitive control \cite{Basar1995}. Here we derived both control types from
the same variational principle.
\end{enumerate}

\section{Conclusion}

In this paper we have proposed a thermodynamic interpretation of bounded rationality based
on a free utility principle. Accordingly, bounded rational agents trade off utility maximization against resource costs measured by the KL divergence with respect to an initial policy. The use of the KL divergence as a cost function for control has been previously proposed to measure deviations from passive dynamics in Markov systems  \cite{Todorov2006,Todorov2009}.
Other methods of statistical physics have been previously proposed as an information-theoretic
approach to interactive learning \cite{Still2009} and to game theory with bounded rational
players \cite{Wolpertt2004}. The contribution of our study is to devise a single axiomatic framework that allows for the treatment of control problems, game-theoretic problems and estimation and learning problems for perfectly rational and bounded rational agents.
In the future it will be interesting to relate the thermodynamic resource costs of
bounded rational agents to more traditional notions of resource
costs in computer science like space and time requirements when computing optimal
actions \cite{Vitanyi2005}.

\bibliographystyle{plain}
\bibliography{bibliography}









\end{document}